\pdfoutput=1

\documentclass[11pt]{article}

\usepackage[]{acl}

\usepackage{times}
\usepackage{latexsym}

\usepackage[T1]{fontenc}

\usepackage[utf8]{inputenc}

\usepackage{microtype}

\usepackage{inconsolata}

\usepackage{hyperref}       
\usepackage{url}            
\usepackage{booktabs}       
\usepackage{amsfonts}       
\usepackage{nicefrac}       
\usepackage{xcolor}
\usepackage{amsmath}         

\usepackage{graphicx}
\usepackage{multirow}
\usepackage{pifont}

\title{Improving Text Embeddings with Large Language Models}

\author{Liang Wang,~Nan Yang,~Xiaolong Huang,\\
~\textbf{Linjun Yang},~\textbf{Rangan Majumder},~\textbf{Furu Wei}\\
Microsoft Corporation \\
\{wangliang,nanya,xiaolhu,yang.linjun,ranganm,fuwei\}@microsoft.com \\}

\begin{document}
\maketitle
\begin{abstract}
    In this paper,
    we introduce a novel and simple method for obtaining high-quality text embeddings
    using only synthetic data and less than $1$k training steps.
    Unlike existing methods that often depend on multi-stage intermediate pre-training
    with billions of weakly-supervised text pairs,
    followed by fine-tuning with a few labeled datasets,
    our method does not require building complex training pipelines
    or relying on manually collected datasets that are often constrained by task diversity and language coverage.
    We leverage proprietary LLMs to generate diverse synthetic data
    for hundreds of thousands of text embedding tasks across $93$ languages.
    We then fine-tune open-source decoder-only LLMs on the synthetic data using standard contrastive loss.
    Experiments demonstrate that our method achieves strong performance on highly competitive text embedding benchmarks
    without using any labeled data.
    Furthermore, when fine-tuned with a mixture of synthetic and labeled data,
    our model sets new state-of-the-art results on the BEIR and MTEB benchmarks.
\end{abstract}

\section{Introduction}
Text embeddings are vector representations of natural language that encode its semantic information.
They are widely used in various natural language processing (NLP) tasks,
such as information retrieval (IR), question answering, semantic textual similarity, bitext mining, item recommendation, etc.
In the field of IR,
the first-stage retrieval often relies on text embeddings to efficiently recall a small set of candidate documents
from a large-scale corpus using approximate nearest neighbor search techniques.
Embedding-based retrieval is also a crucial component of retrieval-augmented generation (RAG)~\citep{lewis2020retrieval},
which is an emerging paradigm that
enables large language models (LLMs) to access dynamic external knowledge without modifying the model parameters.
Source attribution of generated text is another important application of text embeddings~\citep{gao2023enabling}
that can improve the interpretability and trustworthiness of LLMs.

Previous studies have demonstrated that weighted average of pre-trained word embeddings~\citep{pennington2014glove,arora2019simple}
is a strong baseline for measuring semantic similarity.
However,
these methods fail to capture the rich contextual information of natural language.
With the advent of pre-trained language models ~\citep{Devlin2019BERTPO},
Sentence-BERT~\citep{Reimers2019SentenceBERTSE} and SimCSE~\citep{Gao2021SimCSESC} have been proposed to
learn text embeddings by fine-tuning BERT on natural language inference (NLI) datasets.
To further enhance the performance and robustness of text embeddings,
state-of-the-art methods like E5~\citep{wang2022text} and BGE~\citep{xiao2023c}
employ a more complex multi-stage training paradigm that first pre-trains on billions of weakly-supervised text pairs,
and then fine-tunes on several high-quality labeled datasets.

Existing multi-stage approaches suffer from several drawbacks.
Firstly,
they entail a complex multi-stage training pipeline that demands substantial engineering efforts to curate large amounts of relevance pairs.
Secondly,
they rely on manually collected datasets that are often constrained by the diversity of tasks and the coverage of languages.
For instance,
Instructor~\citep{su2022one} is only trained on instructions from $330$ English datasets,
whereas BGE~\citep{xiao2023c} only focuses on high-resource languages such as English and Chinese.
Moreover,
most existing methods employ BERT-style encoders as the backbone,
neglecting the recent advances of training better LLMs and related techniques such as context length extension~\citep{Rozire2023CodeLO}.

In this paper,
we propose a novel method for text embeddings that leverages LLMs to overcome the limitations of existing approaches.
We use proprietary LLMs to generate synthetic data for a diverse range of text embedding tasks in $93$ languages,
covering hundreds of thousands of embedding tasks.
Specifically,
we use a two-step prompting strategy that first prompts the LLMs to brainstorm a pool of candidate tasks,
and then prompts the LLMs to generate data conditioned on a given task from the pool.
To cover various application scenarios,
we design multiple prompt templates for each task type
and combine the generated data from different templates to boost diversity.
For the text embedding models,
we opt for fine-tuning powerful open-source LLMs rather than small BERT-style models.
Since LLMs such as Mistral~\citep{jiang2023mistral} have been extensively pre-trained on web-scale data,
contrastive pre-training that proves to be important for BERT models~\citep{wang2022text}
offers little additional benefit.

We demonstrate that Mistral-7B,
when fine-tuned solely on synthetic data,
attains competitive performance on the BEIR~\citep{Thakur2021BEIRAH} and MTEB~\citep{muennighoff2023mteb} benchmarks.
This is particularly intriguing considering that this setting does not involve any labeled data.
When fine-tuned on a mixture of synthetic and labeled data,
our model achieves new state-of-the-art results,
surpassing previous methods by a significant margin (+$2\%$).
The entire training process requires less than $1$k steps.

Moreover,
we empirically validate that our model can effectively perform personalized passkey retrieval
for inputs up to $32$k tokens by altering the rotation base of the position embeddings,
extending the context length beyond the conventional $512$ token limit.
Regarding its multilinguality,
our model excels on high-resource languages.
However,
for low-resource languages,
there is still room for improvement as current open-source LLMs are not adequately pre-trained on them.

\section{Related Work}
\begin{figure*}[ht]
\begin{center}
 \includegraphics[width=0.9\linewidth]{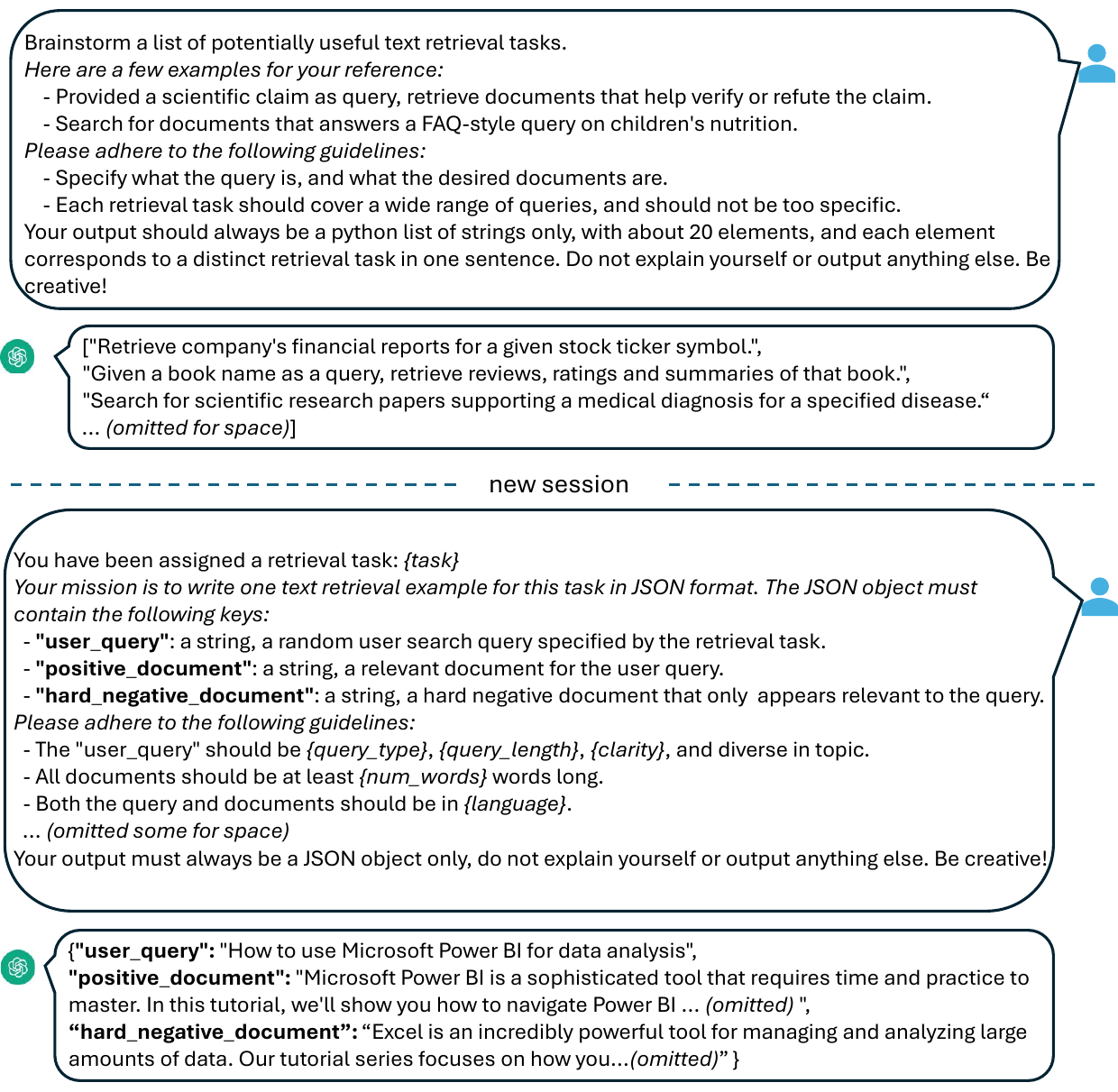}
 \caption{An example two-step prompt template for generating synthetic data with GPT-4.
We first prompt GPT-4 to brainstorm a list of potential retrieval tasks,
and then generate \emph{(query, positive, hard negative)} triplets for each task.
``\emph{\{...\}}'' denotes a placeholder that will be replaced by sampling from a predefined set of values.
Full prompts are available in Appendix ~\ref{sec:app_prompts}.}
 \label{fig:prompt_example}
\end{center}
\end{figure*}

\noindent
\textbf{Text Embeddings }
are continuous low-dimensional representations of text and
have been extensively applied to various downstream tasks such as information retrieval, question answering,
and retrieval-augmented generation (RAG).
Early work on text embeddings includes latent semantic indexing~\citep{deerwester1990indexing}
and weighted average of word embeddings~\citep{Mikolov2013EfficientEO}.
More recent methods exploit supervision from natural language inference~\citep{Bowman2015ALA} and
labeled query-document pairs, such as the MS-MARCO passage ranking dataset~\citep{Campos2016MSMA},
to train text embeddings~\citep{Reimers2019SentenceBERTSE,Conneau2017SupervisedLO,Gao2021SimCSESC}.
However,
labeled data are often limited in terms of task diversity and language coverage.
To address this challenge,
methods like Contriever~\citep{Izacard2021TowardsUD}, OpenAI Embeddings~\citep{Neelakantan2022TextAC}, E5~\citep{wang2022text}, and BGE~\citep{xiao2023c}
adopt a multi-stage training paradigm.
They first pre-train on large-scale weakly-supervised text pairs using contrastive loss
and then fine-tune on small-scale but high-quality datasets.
In this paper,
we demonstrate that it is possible to obtain state-of-the-art text embeddings with single-stage training.
\newline

\noindent
\textbf{Synthetic Data }
Synthetic data generation is a widely studied topic in information retrieval research,
with various methods proposed to enhance retrieval systems with artificially created data.
For instance,
Doc2query~\citep{nogueira2019document}, InPars ~\citep{Bonifacio2022InParsUD}, and Promptagator~\citep{dai2022promptagator}
generate synthetic queries for unlabeled documents,
which are then leveraged for document expansion or model training.
GPL~\citep{Wang2021GPLGP} employs a cross-encoder to produce pseudo-labels
for query-document pairs.
Similarly,
Query2doc~\citep{wang2023query2doc} generates pseudo-documents for query expansion by few-shot prompting LLMs.
Unlike these methods,
our approach does not rely on any unlabeled documents or queries
and thus can generate more diverse synthetic data.

Another related line of work focuses on knowledge distillation from black-box LLMs
by training on synthetic data generated from them.
DINO~\citep{schick2021generating} generates synthetic text pairs for semantic textual similarity.
Unnatural Instructions~\citep{honovich2022unnatural} is a synthetic instruction following dataset
by prompting existing LLMs.
Orca~\citep{Mukherjee2023OrcaPL} and Phi~\citep{Gunasekar2023TextbooksAA} propose to train better small language models
by using high-quality synthetic data from GPT-3.5/4~\citep{OpenAI2023GPT4TR}.
\newline

\noindent
\textbf{Large Language Models }
With the popularization of ChatGPT,
large language models (LLMs) have demonstrated remarkable capabilities in instruction following
and few-shot in-context learning~\citep{NEURIPS2020_1457c0d6}.
However,
the most advanced LLMs such as GPT-4~\citep{OpenAI2023GPT4TR} are proprietary and have little technical details disclosed.
To bridge the gap between proprietary and open-source LLMs,
several notable efforts have been made,
such as LLaMA-2~\citep{touvron2023llama} and Mistral~\citep{jiang2023mistral} models.
A major limitation of LLMs is that they lack awareness of recent events and private knowledge.
This issue can be partly mitigated by augmenting LLMs with information retrieved from external sources,
a technique known as retrieval-augmented generation (RAG).
On the other hand,
LLMs can also serve as foundation models to enhance text embeddings.
RepLLaMA~\citep{Ma2023FineTuningLF} proposes to fine-tune LLaMA-2 with bi-encoder architecture for ad-hoc retrieval.
SGPT~\citep{Muennighoff2022SGPTGS}, GTR~\citep{Ni2021LargeDE}, and Udever~\citep{zhang2023language}
demonstrate the scaling law of text embeddings empirically,
but their performance still falls behind small bidirectional encoders such as E5~\citep{wang2022text} and BGE~\citep{xiao2023c}.
In this paper,
we present a novel approach to train state-of-the-art text embeddings by exploiting the latest advances of LLMs and synthetic data.

\section{Method}

\subsection{Synthetic Data Generation} \label{sec:synthetic_data_generation}
Utilizing synthetic data generated by advanced LLMs such as GPT-4 presents a compelling opportunity,
especially in terms of enhancing diversity across a multitude of tasks and languages.
Such diversity is essential for developing robust text embeddings that can perform well across different tasks,
be it semantic retrieval, textual similarity, or clustering.

To generate diverse synthetic data,
we propose a simple taxonomy that categorizes embedding tasks into several groups,
and then apply different prompt templates to each group.
\newline

\noindent
\textbf{Asymmetric Tasks }
This category comprises tasks where
the query and document are semantically related but are not paraphrases of each other.
Depending on the length of the query and document,
we further divide asymmetric tasks into four subgroups:
short-long match, long-short match, short-short match, and long-long match.
For instance,
short-long match tasks involve a short query and a long document,
which is a typical scenario in commercial search engines.
For each subgroup,
we design a two-step prompt template that first prompts LLMs brainstorm a list of tasks,
and then generates a concrete example conditioned on the task definition.
In Figure ~\ref{fig:prompt_example},
we show an example prompt for the short-long match subgroup.
The full output is available in Table~\ref{tab:app_synthetic_data_examples}.
The outputs from GPT-4 are mostly coherent and of high quality.
In our preliminary experiments,
we also attempted to generate the task definition and query-document pairs using a single prompt,
but the data diversity was not as satisfactory as the proposed two-step approach.
\newline

\noindent
\textbf{Symmetric Tasks }
Symmetric tasks involve queries and documents that have similar semantic meanings but different surface forms.
We examine two application scenarios:
monolingual semantic textual similarity (STS) and bitext retrieval.
We design two distinct prompt templates for each scenario,
tailored to their specific objectives.
Since the task definition is straightforward,
we omit the brainstorming step for symmetric tasks.
\newline

To further boost the diversity of the prompts and thus the synthetic data,
we incorporate several placeholders in each prompt template,
whose values are randomly sampled at runtime.
For example,
in Figure ~\ref{fig:prompt_example},
the value of ``\emph{\{query\_length\}}'' is sampled from the set ``\emph{\{less than 5 words, 5-10 words, at least 10 words\}}''.

To generate multilingual data,
we sample the value of ``\emph{\{language\}}'' from the language list of XLM-R~\citep{conneau2020unsupervised},
giving more weight to high-resource languages.
Any generated data that does not conform to the predefined JSON format
are discarded during the parsing process.
We also remove duplicates based on exact string matching.

\subsection{Training}
Given a relevant query-document pair ($q^+, d^+$),
we first apply the following instruction template to the original query $q^+$ to generate a new one $q^+_\text{inst}$:
\begin{equation} \label{equ:instruction_template}
    q^+_\text{inst}=\text{Instruct: \{task\_definition\}}\ \backslash n\ \text{Query:\ }\{q^+\}
\end{equation}
where ``\emph{\{task\_definition\}}'' is a placeholder for a one-sentence description of the embedding task.
For generated synthetic data,
we use the outputs from the brainstorming step.
For other datasets,
such as MS-MARCO,
we manually craft the task definitions and apply them to all the queries in the dataset.
We do not modify the document side with any instruction prefix.
In this way,
the document index can be prebuilt, and we can customize the task to perform by changing only the query side.

Given a pretrained LLM,
we append an [EOS] token to the end of the query and document,
and then feed them into the LLM to obtain the query and document embeddings ($\mathbf{h}_{q^+_\text{inst}}, \mathbf{h}_{d^+}$)
by taking the last layer [EOS] vector.
To train the embedding model,
we adopt the standard InfoNCE loss $\mathbb{L}$ over the in-batch negatives and hard negatives:
\begin{equation} \label{equ:infonce}
    \min\ \ \mathbb{L} = -\log \frac{\phi(q^+_\text{inst}, d^+)}{\phi(q^+_\text{inst}, d^+) + \displaystyle\sum_{n_i \in \mathbb{N}}(\phi(q^+_\text{inst}, n_i))}
\end{equation}
where $\mathbb{N}$ denotes the set of all negatives,
and $\phi(q, d)$ is a function that computes the matching score between query $q$ and document $d$.
In this paper,
we adopt the temperature-scaled cosine similarity function as follows:
\begin{equation} \label{equ:cosine_similarity}
\phi(q, d) = \text{exp}(\frac{1}{\tau}\cos(\mathbf{h}_q, \mathbf{h}_d))
\end{equation}
$\tau$ is a temperature hyper-parameter,
which is fixed to $0.02$ in our experiments.

\section{Experiments}

\subsection{Statistics of the Synthetic Data}

\begin{figure*}[ht]
\begin{center}
 \includegraphics[width=0.8\linewidth]{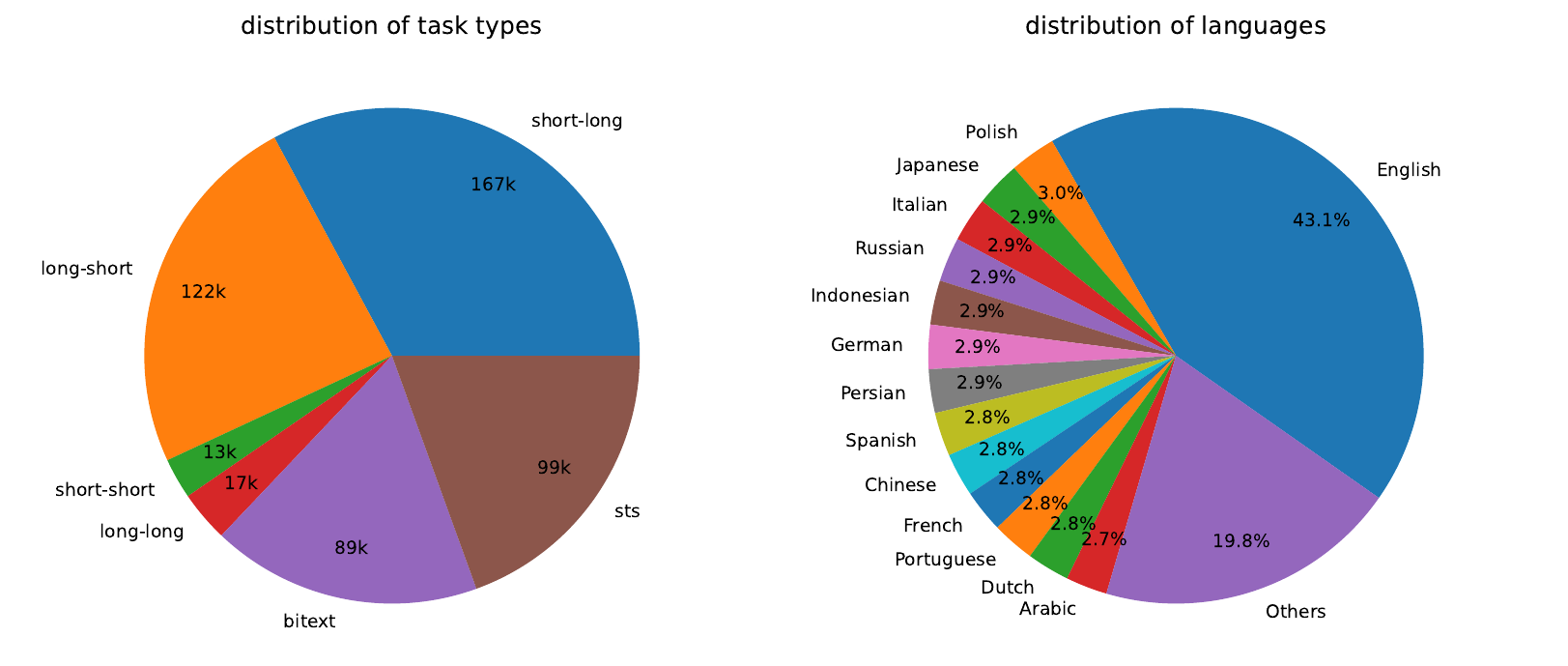}
 \caption{Task type and language statistics of the generated synthetic data (see Section~\ref{sec:synthetic_data_generation} for task type definitions).
The ``Others'' category contains the remaining languages from the XLM-R language list.}
 \label{fig:stat_synthetic_data}
\end{center}
\end{figure*}

\begin{table*}[ht]
\centering
\scalebox{0.95}{\begin{tabular}{lcccccccc}
\hline
\multicolumn{1}{l}{\multirow{2}{*}{\# of datasets $\rightarrow$}} & Class. & Clust. & PairClass. & Rerank & Retr. & STS & Summ. & Avg \\
\multicolumn{1}{l}{}                                & 12     & 11     & 3          & 4      & 15    & 10  & 1    & 56  \\ \hline
\multicolumn{9}{l}{\emph{Unsupervised Models}}    \\ \hline
\multicolumn{1}{l}{Glove~\citep{pennington2014glove}}   &  57.3   &  27.7 & 70.9  & 43.3  & 21.6 & 61.9 & 28.9  & 42.0 \\
\multicolumn{1}{l}{SimCSE$_\text{bert-unsup}$~\citep{Gao2021SimCSESC}}   &  62.5  &  29.0  &  70.3   &  46.5  & 20.3  & 74.3 & 31.2  & 45.5 \\ \hline
\multicolumn{9}{l}{\emph{Supervised Models}}  \\ \hline
\multicolumn{1}{l}{SimCSE$_\text{bert-sup}$~\citep{Gao2021SimCSESC}}  &  67.3  &  33.4  &  73.7   &  47.5  & 21.8  &  79.1    &  23.3    &  48.7 \\
\multicolumn{1}{l}{Contriever~\citep{Izacard2021TowardsUD}}   &   66.7  &  41.1  &  82.5  & 53.1  & 41.9   &  76.5   &  30.4    &  56.0   \\
\multicolumn{1}{l}{GTR$_\text{xxl}$~\citep{Ni2021LargeDE}}   &  67.4  &  42.4   & 86.1   &  56.7   &  48.5  &  78.4   & 30.6  &  59.0 \\
\multicolumn{1}{l}{Sentence-T5$_\text{xxl}$~\citep{ni2022sentence}}   &  73.4   &  43.7  &  85.1    &  56.4  & 42.2  & 82.6  & 30.1  &  59.5 \\
\multicolumn{1}{l}{E5$_\text{large-v2}$~\citep{wang2022text}}     & 75.2  & 44.5  & 86.0  & 56.6  & 50.6  & 82.1 & 30.2  & 62.3   \\
\multicolumn{1}{l}{GTE$_\text{large}$~\citep{li2023towards}}     & 73.3  & 46.8  & 85.0   & 59.1  & 52.2   & 83.4  & 31.7  & 63.1   \\
\multicolumn{1}{l}{BGE$_\text{large-en-v1.5}$~\citep{xiao2023c}}   & 76.0  & 46.1  & 87.1  & 60.0  & 54.3  & 83.1  &  31.6 & 64.2  \\ \hline
\multicolumn{9}{l}{\emph{Ours}}        \\ \hline
\multicolumn{1}{l}{E5$_\text{mistral-7b}$ + full data}  & \textbf{78.5}  & 50.3  & \textbf{88.3}  &  \textbf{60.2}  & \textbf{56.9}  & \textbf{84.6}  & 31.4 & \textbf{66.6} \\
\multicolumn{1}{l}{\ \ w/ synthetic data only}  & 78.2 & \textbf{50.5} & 86.0 & 59.0 & 46.9 & 81.2 & 31.9 & 63.1  \\
\multicolumn{1}{l}{\ \ w/ synthetic + msmarco}  & 78.3 & 49.9 & 87.1 & 59.5 & 52.2 & 81.2 & \textbf{32.7} & 64.5 \\ \hline
\end{tabular}}
\caption{Results on the MTEB benchmark~\citep{muennighoff2023mteb} (56 datasets in the English subset).
The numbers are averaged for each category.
Please refer to Table ~\ref{tab:app_full_results} for the scores per dataset.}
\label{tab:mteb}
\end{table*}

Figure ~\ref{fig:stat_synthetic_data} presents the statistics of our generated synthetic data.
We manage to generate $500$k examples with $150$k unique instructions using Azure OpenAI Service~\footnote{\url{https://oai.azure.com/}},
among which $25\%$ are generated by \emph{GPT-35-Turbo} and others are generated by \emph{GPT-4}.
The total token consumption is about $180$M.
The predominant language is English,
with coverage extending to a total of $93$ languages.
For the bottom $75$ low-resource languages,
there are about $1$k examples per language on average.
Please see Table ~\ref{tab:app_synthetic_data_examples} in the appendix for examples of synthetic data.

In terms of data quality,
we find that a portion of \emph{GPT-35-Turbo} outputs do not strictly follow the guidelines specified in the prompt templates.
Nevertheless,
the overall quality remains acceptable,
and preliminary experiments have demonstrated the benefits of incorporating this data subset.

\subsection{Model Fine-tuning and Evaluation}
\begin{table*}[ht]
\centering
\scalebox{0.9}{\begin{tabular}{lcccccccc}
\hline
     & \multicolumn{4}{c}{High-resource Languages} & \multicolumn{4}{c}{Low-resource Languages} \\ \hline
     & en        & fr        & es       & ru       &  te  &     hi     &   bn       &   sw       \\ \hline
BM25~\citep{Zhang2023MIRACLAM} &     35.1      &    18.3   &    31.9  & 33.4  &   49.4   &     45.8   &      50.8    &   38.3     \\
mDPR~\citep{Zhang2023MIRACLAM} &     39.4    &  43.5  &    47.8   &  40.7 &   35.6  &   38.3  &    44.3      &  29.9   \\
mE5$_\text{base}$~\citep{wang2024multilingual}     &   51.2   & 49.7  &   51.5  & 61.5     &     75.2     &    58.4    &  70.2      &  71.1    \\
mE5$_\text{large}$~\citep{wang2024multilingual}  &   52.9  &  54.5    &  \textbf{52.9}  & 67.4   &  \textbf{84.6}   &  \textbf{62.0}   &  \textbf{75.9}  &  \textbf{74.9}  \\ \hline
E5$_\text{mistral-7b}$ + full data  &  \textbf{57.3}    &  \textbf{55.2}   & 52.2  & \textbf{67.7}   &  73.9     & 52.1    &   70.3   & 68.4    \\ \hline
\end{tabular}}
\caption{nDCG@10 on the dev set of the MIRACL dataset for both high-resource and low-resource languages.
We select the $4$ high-resource languages and the $4$ low-resource languages
according to the number of candidate documents.
The numbers for BM25 and mDPR come from ~\citet{Zhang2023MIRACLAM}.
For the complete results on all $16$ languages, please see Table~\ref{tab:app_full_miracl_ndcg}.}
\label{tab:miracl_high_low_languages}
\end{table*}

\begin{figure*}[ht]
\begin{center}
 \includegraphics[width=0.95\linewidth]{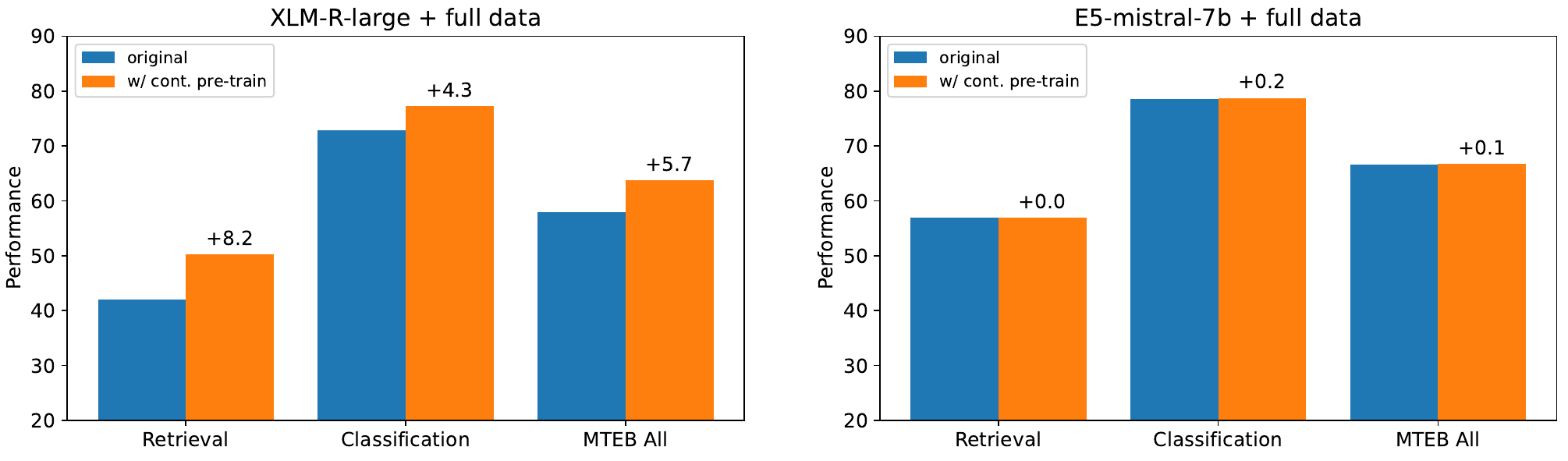}
 \caption{Effects of contrastive pre-training.
Detailed numbers are in Appendix Table~\ref{tab:app_ablation_pretrain}.}
 \label{fig:ablation_cont_pretrain}
\end{center}
\end{figure*}

The pretrained Mistral-7b~\citep{jiang2023mistral} checkpoint is fine-tuned for $1$ epoch using the loss in Equation ~\ref{equ:infonce}.
We follow the training recipe from RankLLaMA~\citep{Ma2023FineTuningLF}
and utilize LoRA~\citep{hu2021lora} with rank $16$.
To further reduce GPU memory requirement,
techniques including gradient checkpointing, mixed precision training, and DeepSpeed ZeRO-3 are applied.

For the training data,
we utilize both the generated synthetic data
and a collection of $13$ public datasets,
yielding approximately $1.8$M examples after sampling.
More details are available in Appendix ~\ref{sec:app_implementation}.
To provide a fair comparison with some previous work,
we also report results when the only labeled supervision is the MS-MARCO passage ranking~\citep{Campos2016MSMA} dataset.

We evaluate the trained model on the MTEB benchmark~\citep{muennighoff2023mteb}.
Note that the retrieval category in MTEB corresponds to the $15$ publicly available datasets
in the BEIR benchmark~\citep{Thakur2021BEIRAH}.
Evaluation of one model takes about $3$ days on $8$ V100 GPUs due to the need to encode
a large number of documents.
Although our model can accommodate sequence length beyond $512$,
we only evaluate on the first $512$ tokens for efficiency.
Official metrics are reported for each category.
For more details about the evaluation protocol,
please refer to the original papers~\citep{muennighoff2023mteb,Thakur2021BEIRAH}.

\subsection{Main Results}

In Table~\ref{tab:mteb},
our model ``E5$_\text{mistral-7b}$ + full data'' attains the highest average score on the MTEB benchmark,
outperforming the previous state-of-the-art model by $2.4$ points.
In the ``w/ synthetic data only'' setting,
no labeled data is used for training,
and yet the performance remains quite competitive.
We posit that generative language modeling and text embeddings are the two sides of the same coin,
with both tasks requiring the model to have a deep understanding of the natural language.
Given an embedding task definition,
a truly robust LLM should be able to generate training data on its own and
then be transformed into an embedding model through light-weight fine-tuning.
Our experiments shed light on the potential of this direction,
and more research is needed to fully explore it.

\begin{table}[ht]
\centering
\scalebox{0.95}{\begin{tabular}{lcc}
\hline
Model & BEIR & MTEB \\ \hline
OpenAI text-embedding-3-large  &  55.4   & 64.6  \\
Cohere-embed-english-v3.0  &   55.0    & 64.5  \\
voyage-lite-01-instruct   &    55.6   & 64.5  \\
UAE-Large-V1  &   54.7    &   64.6  \\ \hline
E5$_\text{mistral-7b}$ + full data      &   \textbf{56.9}   &  \textbf{66.6}   \\ \hline
\end{tabular}}
\caption{Comparison with commercial models and the model that tops the MTEB leaderboard (as of 2023-12-22)~\citep{li2023angle}.
``BEIR'' is the average nDCG@10 score over $15$ public datasets in the BEIR benchmark~\citep{Thakur2021BEIRAH}.
``MTEB'' is the average score over $56$ datasets in the English subset of the MTEB benchmark~\citep{muennighoff2023mteb}.
For the commercial models listed here,
little details are available on their model architectures and training data.}
\label{tab:comparison_commercial}
\end{table}

\begin{figure*}[ht]
\begin{center}
 \includegraphics[width=1.0\linewidth]{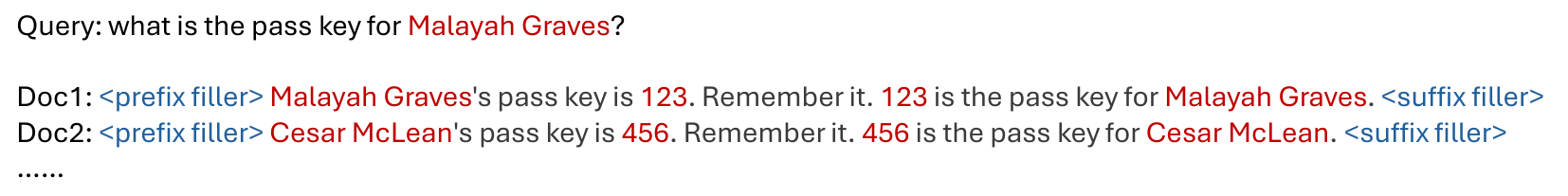}
 \caption{Illustration of the \emph{personalized passkey retrieval} task adapted from ~\citet{mohtashami2023landmark}.
The ``\emph{<prefix filler>}'' and ``\emph{<suffix filler>}'' are repeats of
``\emph{The grass is green. The sky is blue. The sun is yellow. Here we go. There and back again.}''
In addition,
each document has a unique person name and a random passkey inserted at a random position.
The task is to retrieve the document that contains the given person's passkey from $100$ candidates.}
 \label{fig:task_passkey}
\end{center}
\end{figure*}

\begin{figure*}[ht]
\begin{center}
 \includegraphics[width=1.0\linewidth]{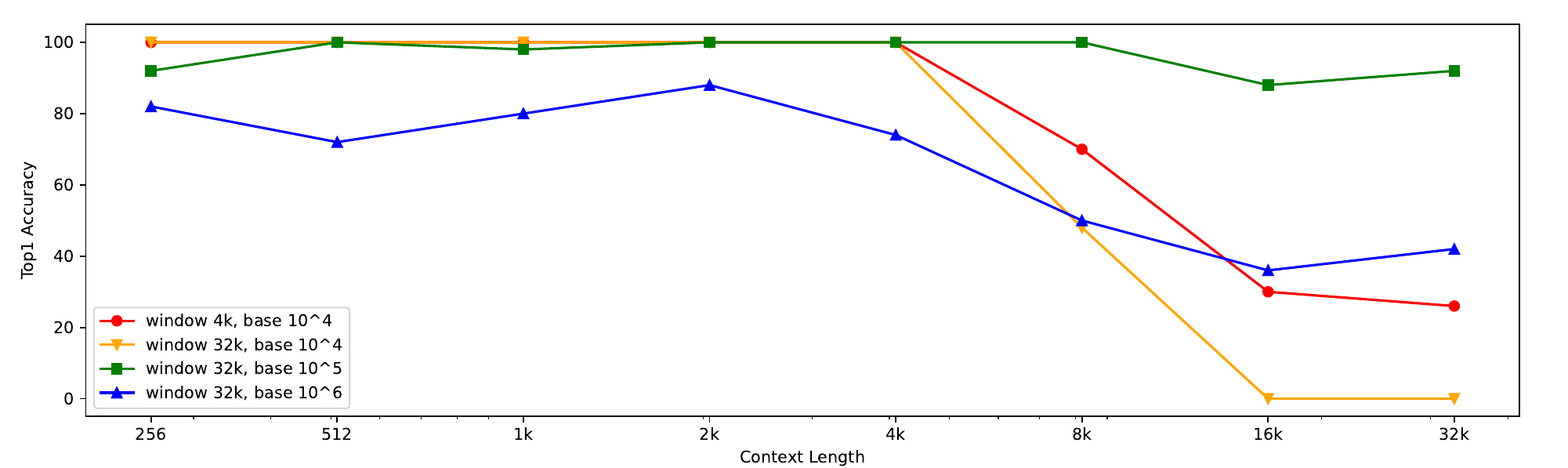}
 \caption{Accuracy of personalized passkey retrieval as a function of input context length.
For each context length,
we randomly generate $50$ queries and compute the top-1 accuracy.}
 \label{fig:passkey}
\end{center}
\end{figure*}

In Table~\ref{tab:comparison_commercial},
we also present a comparison with several commercial text embedding models.
However,
due to the lack of transparency and documentation about these models,
a fair comparison is not feasible.
We focus especially on the retrieval performance on the BEIR benchmark,
since retrieval-augmented generation is an emerging technique to enhance LLM with external knowledge and proprietary data.
As Table~\ref{tab:comparison_commercial} shows,
our model outperforms the current commercial models by a significant margin.

\subsection{Multilingual Retrieval}
To assess the multilingual capabilities of our model,
we conduct an evaluation on the MIRACL dataset~\citep{Zhang2023MIRACLAM},
which comprises human-annotated queries and relevance judgments across $18$ languages.
The validation set contains labels for $16$ languages.
As shown in Table~\ref{tab:miracl_high_low_languages},
our model surpasses mE5$_\text{large}$ on high-resource languages, notably on English.
Nevertheless,
for low-resource languages,
our model remains suboptimal compared to mE5$_\text{base}$.
We attribute this to the fact that Mistral-7B is predominantly pre-trained on English data,
and we anticipate that future multilingual LLMs will leverage our method to bridge this gap.

\begin{table}[ht]
\centering
\scalebox{1.0}{\begin{tabular}{lcc}
\hline
 & \begin{tabular}[c]{@{}c@{}}BUCC 2018\\ 4 langs\end{tabular} & \begin{tabular}[c]{@{}c@{}}Tatoeba\\ 112 langs\end{tabular} \\ \hline
mContriever & 93.7 & 37.7 \\
LaBSE & 98.8 & \textbf{81.1} \\
mE5$_\text{base}$ & 98.1 &  68.1 \\
mE5$_\text{large}$ & 98.6 &  75.7 \\ \hline
E5$_\text{mistral-7b}$ & \textbf{98.9} & 70.1 \\ \hline
\end{tabular}}
\caption{Bitext mining results.
BUCC 2018~\citep{zweigenbaum2018overview} contains $4$ high-resource languages.
Tatoeba~\citep{Artetxe2019MassivelyMS} consists of $112$ English-centric language pairs.}
\label{tab:bitext}
\end{table}

To evaluate our model's cross-lingual retrieval capability,
we report Bitext mining results in Table ~\ref{tab:bitext}.
For baselines including mContriever~\citep{Izacard2021TowardsUD},
LaBSE~\citep{feng2022language}, and mE5~\citep{wang2024multilingual},
we evaluate the results using publicly available checkpoints.
Our observations indicate that,
similar to the MIRACL retrieval,
E5$_\text{mistral-7b}$ excels in bitext mining for high-resource languages only.

\begin{table*}[ht]
\centering
\scalebox{0.95}{\begin{tabular}{lcccccccc}
\hline
\multicolumn{1}{l}{Datasets} & Class. & Clust. & PairClass. & Rerank & Retr. & STS & Summ. & Avg \\ \hline
\multicolumn{1}{l}{E5$_\text{mistral-7b}$}  & 78.3 & 49.9 & 87.1 & \textbf{59.5} & 52.2 & 81.2 & \textbf{32.7} & 64.5 \\ \hline
\multicolumn{1}{l}{\ \ w/ LLaMA-2 7b init.}  & 76.2 & 48.1 & 85.1 & 58.9 & 49.6 & 81.2 & 30.8 & 62.9$^\text{-1.6}$ \\
\multicolumn{1}{l}{\ \ w/ msmarco data only}  & 71.6 & 47.1 & 86.1 & 58.8 & 54.4 & 79.5 & 31.7 & 62.7$^\text{-1.8}$ \\ \hline
\multicolumn{9}{l}{\emph{pooling type}}        \\
\multicolumn{1}{l}{\ \ w/ mean pool}  & 77.0 & 48.9 & 86.1 & 59.2 & 52.4 & \textbf{81.4} & 30.8 & 64.1$^\text{-0.4}$ \\
\multicolumn{1}{l}{\ \ w/ weighted mean}  & 77.0 & 49.0 & 86.1 & 59.2 & 52.0 & \textbf{81.4} & 30.2 & 64.0$^\text{-0.5}$ \\ \hline
\multicolumn{9}{l}{\emph{LoRA rank}}        \\
\multicolumn{1}{l}{\ \ w/ r=$8$}  & \textbf{78.4} & \textbf{50.3} & 87.1 & 59.3 & \textbf{53.0} & 81.0 & 31.7 & \textbf{64.8}$^\text{+0.3}$  \\
\multicolumn{1}{l}{\ \ w/ r=$32$}  & \textbf{78.4} & \textbf{50.3} & \textbf{87.4} & \textbf{59.5 }& 52.2 & 81.2 & 30.6 & 64.6$^\text{+0.1}$ \\ \hline
\multicolumn{9}{l}{\emph{instruction type}}        \\
\multicolumn{1}{l}{\ \ w/o instruction}  & 72.3 & 47.1 & 82.6 & 56.3 & 48.2 & 76.7 & 30.7 & 60.3$^\text{-4.2}$ \\
\multicolumn{1}{l}{\ \ w/ task type prefix}  & 71.1 & 46.5 & 79.7 & 54.0 & 52.7 & 73.8 & 30.0 & 60.3$^\text{-4.2}$ \\ \hline
\end{tabular}}
\caption{Results on the MTEB benchmark with various hyperparameters.
The first row corresponds to the default setting,
which employs last-token pooling, LoRA rank $16$, and natural language instructions.
Unless otherwise stated, all models are trained on the synthetic and MS-MARCO passage ranking data.}
\label{tab:ablation_hyperparams}
\end{table*}

\section{Analysis}

\subsection{Is Contrastive Pre-training Necessary?}
Weakly-supervised contrastive pre-training is one of the key factors
behind the success of existing text embedding models.
For instance,
Contriever~\citep{Izacard2021TowardsUD} treats random cropped spans as positive pairs for pre-training,
while E5~\citep{wang2022text} and BGE~\citep{xiao2023c} collect and filter text pairs from various sources.

This section re-evaluates the necessity of contrastive pre-training for LLMs,
particularly those that have been pre-trained on trillions of tokens.
Figure~\ref{fig:ablation_cont_pretrain} shows that contrastive pre-training benefits XLM-R$_\text{large}$,
enhancing its retrieval performance by $8.2$ points when fine-tuned on the same data,
which aligns with prior findings.
However, for Mistral-7B based models,
contrastive pre-training has negligible impact on the model quality.
This implies that extensive auto-regressive pre-training enables LLMs to acquire good text representations,
and only minimal fine-tuning is required to transform them into effective embedding models.

\subsection{Extending to Long Text Embeddings}

Existing evaluation datasets for text embedding models are typically short,
to evaluate the long-context capability of our model,
we introduce a novel synthetic task called \emph{personalized passkey retrieval},
which is illustrated in Figure~\ref{fig:task_passkey}.
This task requires encoding the passkey information in a long context into the embeddings.
We compare the performance of different variants by changing the sliding window size
and the RoPE rotation base~\citep{su2024roformer} in Figure ~\ref{fig:passkey}.
The results show that the default configuration with $4$k sliding window attains $100\%$ accuracy within $4$k tokens,
but the accuracy deteriorates quickly as the context length grows.
Naively extending the sliding window size to $32$k results in worse performance.
By changing the RoPE rotation base to $10^5$,
the model can achieve over $90\%$ accuracy within $32$k tokens.
However,
this entails a minor trade-off in performance for shorter contexts.
A potential avenue for future research is to efficiently adapt the model to longer contexts
through lightweight post-training~\citep{zhu2023pose}.

\subsection{Analysis of Training Hyperparameters}

Table~\ref{tab:ablation_hyperparams} presents the results under different configurations.
We notice that the Mistral-7B initialization holds an advantage over LLaMA-2 7B,
in line with the findings from Mistral-7B technical report~\citep{jiang2023mistral}.
The choice of pooling types and LoRA ranks does not affect the overall performance substantially,
hence we adhere to the default setting despite the marginal superiority of LoRA rank $8$.
On the other hand,
the way of adding instructions has a considerable impact on the performance.
We conjecture that natural language instructions better inform the model
regarding the embedding task at hand,
and thus enable the model to generate more discriminative embeddings.
Our framework also provides a way to customize the behavior of text embeddings through instructions
without the need to fine-tune the model or re-build document index.

\section{Conclusion}
This paper shows that the quality of text embeddings can be substantially enhanced by exploiting LLMs.
We prompt proprietary LLMs such as GPT-4 to generate diverse synthetic data with instructions in many languages.
Combined with the strong language understanding capability of the Mistral model,
we establish new state-of-the-art results for nearly all task categories on the competitive MTEB benchmark.
The training process is much more streamlined and efficient than existing multi-stage approaches,
thereby obviating the need for intermediate pre-training.

For future work,
we aim to further improve the multilingual performance of our model
and explore the possibility of using open-source LLMs to generate synthetic data.

\section*{Limitations}
In comparison to the mainstream BERT-style encoders,
the employment of LLMs, such as Mistral-7B,
for text embeddings results in a significantly increased inference cost.
The development of more advanced GPUs and better kernel implementations
may enhance the efficiency of the inference process.
With regards to storage cost,
our model is comparatively more expensive,
with embeddings of $4096$ dimensions.
Early successes in reducing embedding dimensions while maintaining competitive performance have been demonstrated
through techniques such as Matryoshka representation learning~\citep{Kusupati2022MatryoshkaRL}.

For synthetic data generation,
we rely on manual prompt engineering to elicit high-quality outputs from proprietary LLMs.
Automatic prompt optimization presents a promising avenue for improving the quality of synthetic data.

\section*{Acknowledgements}
We would like to thank anonymous reviewers for their valuable comments,
and ACL 2024 and ACL Rolling Review organizers for their efforts.
Opinions expressed in this paper are solely those of the authors and do not represent the views of their employers.

\bibliography{custom}

\appendix

\section{Implementation Details} \label{sec:app_implementation}

\begin{table*}[ht]
\centering
\scalebox{0.8}{\begin{tabular}{lcccccccccc}
\hline
& \multicolumn{5}{c}{nDCG@10} & \multicolumn{5}{c}{Recall@100} \\ \hline
& BM25 & mDPR & mE5$_\text{base}$ & mE5$_\text{large}$ & E5$_\text{mistral-7b}$ full & BM25 & mDPR & mE5$_\text{base}$ & mE5$_\text{large}$ & E5$_\text{mistral-7b}$ full \\ \hline
ar & 48.1 & 49.9 & 71.6 & 76.0 & 73.3 & 88.9 & 84.1 & 95.9 & 97.3 & 96.0 \\
bn & 50.8 & 44.3 & 70.2 & 75.9 & 70.3 & 90.9 & 81.9 & 96.6 & 98.2 & 96.0 \\
en & 35.1 & 39.4 & 51.2 & 52.9 & 57.3 & 81.9 & 76.8 & 86.4 & 87.6 &  90.2 \\
es & 31.9 & 47.8 & 51.5 & 52.9 & 52.2 & 70.2 & 86.4 & 88.6 & 89.1 & 87.5  \\
fa & 33.3 & 48.0 & 57.4 & 59.0 & 52.1 & 73.1 & 89.8 & 91.2 & 92.9 & 88.0  \\
fi & 55.1 & 47.2 & 74.4 & 77.8 & 74.7 & 89.1 & 78.8 & 96.9 & 98.1 & 96.7  \\
fr & 18.3 & 43.5 & 49.7 & 54.5 & 55.2 & 65.3 & 91.5 & 90.0 & 90.6 & 92.8  \\
hi & 45.8 & 38.3 & 58.4 & 62.0 & 52.1 & 86.8 & 77.6 & 92.6 & 93.9 & 89.9  \\
id & 44.9 & 27.2 & 51.1 & 52.9 & 52.7 & 90.4 & 57.3 & 87.4 & 87.9 &  88.4 \\
ja & 36.9 & 43.9 & 64.7 & 70.6 & 66.8 & 80.5 & 82.5 & 96.0 & 97.1 & 95.1  \\
ko & 41.9 & 41.9 & 62.2 & 66.5 & 61.8 & 78.3 & 73.7 & 91.6 & 93.4 & 89.4  \\
ru & 33.4 & 40.7 & 61.5 & 67.4 & 67.7 & 66.1 & 79.7 & 92.7 & 95.5 & 95.0  \\
sw & 38.3 & 29.9 & 71.1 & 74.9 & 68.4 & 70.1 & 61.6 & 95.6 & 96.7 &  95.5 \\
te & 49.4 & 35.6 & 75.2 & 84.6 & 73.9 & 83.1 & 76.2 & 98.0 & 99.2  &  95.1 \\
th & 48.4 & 35.8 & 75.2 & 80.2 & 74.0 & 88.7 & 67.8 & 98.0 & 98.9 &  96.5 \\
zh & 18.0 & 51.2 & 51.5 & 56.0 & 54.0 & 56.0 & 94.4 & 92.1 & 93.3 & 90.1  \\ \hline
Avg & 39.3 & 41.5 & 62.3 & 66.5 & 62.9 & 78.7 & 78.8 & 93.1 & 94.3 & 92.6  \\ \hline
\end{tabular}}
\caption{nDCG@10 and Recall@100 on the dev set of the MIRACL dataset for all $16$ languages.}
\label{tab:app_full_miracl_ndcg}
\end{table*}

\begin{table*}[ht]
\centering
\scalebox{0.95}{\begin{tabular}{lcccccccc}
\hline
\multicolumn{1}{l}{Datasets} & Class. & Clust. & PairClass. & Rerank & Retr. & STS & Summ. & Avg \\ \hline
\multicolumn{1}{l}{XLM-R$_\text{large}$ + full data}  & 72.9 & 38.7 & 84.5 & 53.8 & 42.0 & 82.3 & 29.7  & 58.0 \\
\multicolumn{1}{l}{\ \ w/ cont. pre-train}  & 77.2 & 47.3 & 85.5 & 58.6 & 50.2 & 84.4 & 30.7 & 63.7  \\ \hline
\multicolumn{1}{l}{E5$_\text{mistral-7b}$ + full data}  & 78.5  & 50.3  & 88.3  &  60.2  & 56.9  & 84.6  & 31.4 & 66.6 \\
\multicolumn{1}{l}{\ \ w/ cont. pre-train}  & 78.7 & 50.1 & 87.7 & 60.9 & 56.9 & 84.9 & 30.2 & 66.7 \\ \hline
\end{tabular}}
\caption{Detailed results for the effects of contrastive pre-training.
For the ``E5$_\text{mistral-7b}$ w/ cont. pre-train'' setting,
we pre-train Mistral-7B following the mE5 recipe for $10$k steps.}
\label{tab:app_ablation_pretrain}
\end{table*}

\noindent
\textbf{Baseline Models }
For results with mE5$_\text{base}$ and mE5$_\text{large}$,
we use the public checkpoints available
at \url{https://huggingface.co/intfloat/multilingual-e5-base} and \url{https://huggingface.co/intfloat/multilingual-e5-large} respectively.
For experiments in Table~\ref{tab:ablation_hyperparams},
we follow the SGPT~\citep{Muennighoff2022SGPTGS} paper for the implementation of weighted mean pooling.
For the ``w/ task type prefix'' setting,
we prepend ``classify: '' for the long-short matching subgroup,
and ``query: '' for other asymmetric tasks.
No prefix is added for symmetric tasks.
\newline

\noindent
\textbf{Training Data }
For the ``E5$_\text{mistral-7b}$ + full data'' setting,
our training data comprises generated synthetic data, ELI5~\citep{fan2019eli5}(sample ratio $0.1$), HotpotQA~\citep{yang2018hotpotqa}, FEVER~\citep{Thorne2018FEVERAL},
MIRACL~\citep{Zhang2023MIRACLAM},
MSMARCO passage ranking (sample ratio $0.5$) and document ranking (sample ratio $0.2$)~\citep{Campos2016MSMA},
NQ~\citep{Karpukhin2020DensePR},
NLI~\citep{Gao2021SimCSESC}, SQuAD~\citep{Karpukhin2020DensePR}, TriviaQA~\citep{Karpukhin2020DensePR},
Quora Duplicate Questions~\citep{quora-question-pairs}(sample ratio $0.1$), MrTyDi~\citep{zhang2021mr},
DuReader~\citep{qiu2022dureader}, and T2Ranking~\citep{xie2023t2ranking}(sample ratio $0.5$) datasets.
We only include the training set of each dataset.
For the datasets without hard negatives,
we use mE5$_\text{base}$ to mine top $100$ hard negatives.
After sampling,
we obtain approximately $1.8$ million examples.
The entire training process takes fewer than $1$k steps to complete.
\newline

\noindent
\textbf{Hyperparameters for Fine-tuning }
When fine-tuning Mistral-7b~\footnote{\url{https://huggingface.co/mistralai/Mistral-7B-v0.1}},
the batch size is set to $2048$ and the learning rate is $10^{-4}$ with $100$ step warmup and linear decay.
The weight decay is $0.1$.
We add $1$ hard negative for each query-document pair.
The fine-tuning process takes roughly $18$ hours on $32$ V100 GPUs with a maximum sequence length $512$.
We add LoRA adapters to all linear layers,
resulting in a total of $42$M trainable parameters.
Our implementation is based on the HuggingFace PEFT library at \url{https://github.com/huggingface/peft}.
\newline

\noindent
\textbf{Artifacts }
The model and dataset release information is available at \url{https://github.com/microsoft/unilm/tree/master/e5}.
We release our trained models and evaluation scripts to facilitate reproducibility and further research.

\section{Test Set Contamination Analysis} \label{sec:app_test_set_contamination}
To assess the test set contamination on all the datasets in the MTEB benchmark,
we perform a string match based analysis between the test set and our training set,
disregarding differences in character case and spacing.
We categorize the train-test overlaps into three types:
\begin{itemize}
    \item \textbf{Low entropy texts. } These are texts such as ``\emph{i need a coffee}'' and ``\emph{what does that mean}'',
which are not considered as contamination because they are common expressions that can occur in various contexts.
    \item \textbf{Question overlap. } We identify $4$ test set questions in the DBPedia dataset that also appear in the TriviaQA training set.
Given that they constitute a minor portion of the test set, their impact on the overall performance is insignificant.
    \item \textbf{Retrieval corpus overlap. } Several retrieval datasets share the same retrieval corpus.
For instance, the DBPedia, NQ, and TriviaQA datasets all use Wikipedia passages, even though their query sets are different.
This is a standard evaluation practice in the field of information retrieval,
and we do not regard it as contamination.
\end{itemize}
In summary,
we did not detect substantial contamination risks that could alter the main findings of this paper.

Another aspect to consider is the possibility of test set contamination in the training data of Mistral-7B and GPT-4.
However,
since the training data of these models is not publicly accessible,
it is challenging to estimate the degree of such contamination.
Given their widespread use in the research community,
we believe it is still a valid comparison if other works also employ these models.

\begin{table*}[ht]
\centering
\small{\begin{tabular}{l}
\hline
\begin{tabular}[c]{@{}p{0.97\linewidth}@{}}Brainstorm a list of potentially useful text retrieval tasks.\\ \\ Here are a few examples for your reference:\\ - Retrieve relevant documents for a short keyword web search query that asks for weather information.\\ - Search for documents that answers a FAQ-style query on children's nutrition.\\ \\ Please adhere to the following guidelines:\\ - Specify what the query is, and what the desired documents are.\\ - Each retrieval task should cover a wide range of queries, and should not be too specific.\\ \\ Your output must always be a python list of strings only, with about 20 elements, and each element corresponds to a distinct retrieval task in one sentence. Do not explain yourself or output anything else. Be creative!\end{tabular} \\ \hline \hline
\begin{tabular}[c]{@{}p{0.97\linewidth}@{}}You have been assigned a retrieval task: \{task\}\\ \\ Your mission is to write one text retrieval example for this task in JSON format. The JSON object must contain the following keys:\\ - "user\_query": a string, a random user search query specified by the retrieval task.\\ - "positive\_document": a string, a relevant document for the user query.\\ - "hard\_negative\_document": a string, a hard negative document that only appears relevant to the query.\\ \\ Please adhere to the following guidelines:\\ - The "user\_query" should be \{query\_type\}, \{query\_length\}, \{clarity\}, and diverse in topic.\\ - All documents must be created independent of the query. Avoid copying the query verbatim. It’s acceptable if some parts of the "positive\_document" are not topically related to the query.\\ - All documents should be at least \{num\_words\} words long.\\ - The "hard\_negative\_document" contains some useful information, but it should be less useful or comprehensive compared to the "positive\_document".\\ - Both the query and documents should be in \{language\}.\\ - Do not provide any explanation in any document on why it is relevant or not relevant to the query.\\ - Both the query and documents require \{difficulty\} level education to understand.\\ \\ Your output must always be a JSON object only, do not explain yourself or output anything else. Be creative!\end{tabular} \\ \hline
\end{tabular}}
\caption{Prompt template for the short-long matching subgroup.
For placeholders,
``\emph{\{query\_type\}}'' $\in$ \{extremely long-tail, long-tail, common\},
``\emph{\{query\_length\}}'' $\in$ \{less than 5 words, 5 to 15 words, at least 10 words\},
``\emph{\{difficulty\}}'' $\in$ \{high school, college, PhD\},
``\emph{\{clarity\}}'' $\in$ \{clear, understandable with some effort, ambiguous\},
``\emph{\{num\_words\}}'' $\in$ \{50, 100, 200, 300, 400, 500\}.}
\label{tab:app_short_long}
\end{table*}

\begin{table*}[ht]
\centering
\small{\begin{tabular}{l}
\hline
\begin{tabular}[c]{@{}p{0.97\linewidth}@{}}Brainstorm a list of potentially useful text classification tasks.\\ \\ Please adhere to the following guidelines:\\ - Tasks should cover a diverse range of domains and task types.\\ \\ Your output must always be a python list of strings only, with about 20 elements, and each element corresponds to a distinct text classification task in one sentence. Do not explain yourself or output anything else. Be creative!\end{tabular} \\ \hline \hline
\begin{tabular}[c]{@{}p{0.97\linewidth}@{}}You have been assigned a text classification task: \{task\}\\ \\ Your mission is to write one text classification example for this task in JSON format. The JSON object must contain the following keys:\\ - "input\_text": a string, the input text specified by the classification task.\\ - "label": a string, the correct label of the input text.\\ - "misleading\_label": a string, an incorrect label that is related to the task.  \\ \\ Please adhere to the following guidelines:\\ - The "input\_text" should be \{num\_words\} words and diverse in expression.\\ - The "misleading\_label" must be a valid label for the given task, but not as appropriate as the "label" for the "input\_text".\\ - The values for all fields should be in \{language\}.\\ - Avoid including the values of the "label" and "misleading\_label" fields in the "input\_text", that would make the task too easy.\\ - The "input\_text" is \{clarity\} and requires \{difficulty\} level education to comprehend.\\ \\ Your output must always be a JSON object only, do not explain yourself or output anything else. Be creative!\end{tabular} \\ \hline
\end{tabular}}
\caption{Prompt template for the long-short matching subgroup.
For placeholders,
``\emph{\{num\_words\}}'' $\in$ \{"less than 10", "at least 10", "at least 50", "at least 100", "at least 200"\},
``\emph{\{difficulty\}}'' $\in$ \{high school, college, PhD\},
``\emph{\{clarity\}}'' $\in$ \{clear, understandable with some effort, ambiguous\}.}
\label{tab:app_long_short}
\end{table*}

\begin{table*}[ht]
\centering
\small{\begin{tabular}{l}
\hline
\begin{tabular}[c]{@{}p{0.97\linewidth}@{}}Brainstorm a list of text matching tasks where both the queries and the groundtruth documents are very short (one or two sentences, even a short phrase).  \\ \\ Here are a few examples:\\ - Given a scientific paper title, retrieve the title of papers that cite the given paper.\\ - Match a word with its definition.\\ - Provided a notable person's name, identify their occupation or achievement.\\ \\ Your output must always be a python list of strings only, with about 20 elements, and each element corresponds to a distinct task in one sentence. Do not explain yourself or output anything else. Be creative!\end{tabular} \\ \hline \hline
\begin{tabular}[c]{@{}p{0.97\linewidth}@{}}You have been assigned a text matching task: \{task\}\\ \\ Your mission is to write one example for this task in JSON format. The JSON object must contain the following keys:\\ - "input": a string, a random input specified by the task.\\ - "positive\_document": a string, a relevant document for the "input" according to the task.\\ \\ Please adhere to the following guidelines:\\ - The values of all fields should be in \{language\}.\\ - Both the "input" and "positive\_document" should be very short (a sentence or a phrase), avoid substantial word overlaps, otherwise the task would be too easy.\\ - The "input" and "positive\_document" should be independent of each other.\\ \\ Your output must always be a JSON object only, do not explain yourself or output anything else. Be creative!\end{tabular} \\ \hline
\end{tabular}}
\caption{Prompt template for the short-short matching subgroup.
We do not generate negative documents as the matching task is already reasonably difficult.}
\label{tab:app_short_short}
\end{table*}

\begin{table*}[ht]
\centering
\small{\begin{tabular}{l}
\hline
\begin{tabular}[c]{@{}p{0.97\linewidth}@{}}Brainstorm a list of text matching tasks where the queries are long documents.  \\   \\ Here are a few examples:  \\ - Given a document that supports a debatable argument, find another document that contains opposite arguments.  \\ - Provided a lengthy business proposal, retrieve competitive business strategies in the same industry.\\ \\ Your output must always be a python list of strings only, with about 20 elements, and each element corresponds to a distinct task in one sentence. Do not explain yourself or output anything else. Be creative!\end{tabular} \\ \hline \hline
\begin{tabular}[c]{@{}p{0.97\linewidth}@{}}You have been assigned a text matching task: \{task\}\\ \\ Your mission is to write one example for this task in JSON format. The JSON object must contain the following keys:\\ - "input": a string, a random input specified by the task.\\ - "positive\_document": a string, a relevant document for the "input" according to the task.\\ \\ Please adhere to the following guidelines:\\ - The values of all fields should be in \{language\}.\\ - Both the "input" and "positive\_document" should be long documents (at least 300 words), avoid substantial word overlaps, otherwise the task would be too easy.\\ - The "input" and "positive\_document" should be independent of each other.\\ \\ Your output must always be a JSON object only, do not explain yourself or output anything else. Be creative!\end{tabular} \\ \hline
\end{tabular}}
\caption{Prompt template for the long-long matching subgroup.
We do not generate negative documents for API latency reasons.}
\label{tab:app_long_long}
\end{table*}

\begin{table*}[ht]
\centering
\small{\begin{tabular}{l}
\hline
\begin{tabular}[c]{@{}p{0.97\linewidth}@{}}Write a \{unit\} triple with varying semantic similarity scores in JSON format. The semantic similarity score ranges from 1 to 5, with 1 denotes least similar and 5 denotes most similar.\\ \\ Please adhere to the following guidelines:\\ - The keys in JSON are "S1", "S2", and "S3", the values are all strings in \{language\}, do not add any other keys.\\ - There should be some word overlaps between all three \{unit\}s.\\ - The similarity score between S1 and S2 should be \{high\_score\}.\\ - The similarity score between S1 and S3 should be \{low\_score\}.\\ - The \{unit\}s require \{difficulty\} level education to understand and should be diverse in terms of topic and length.\\ \\ Your output must always be a JSON object only with three keys "S1", "S2" and "S3", do not explain yourself or output anything else. Be creative!\end{tabular} \\ \hline
\end{tabular}}
\caption{Prompt template for monolingual STS.
For placeholders,
``\emph{\{high\_score\}}'' $\in$ \{4, 4.5, 5\},
``\emph{\{low\_score\}}'' $\in$ \{2.5, 3, 3.5\},
``\emph{\{unit\}}'' $\in$ \{sentence, phrase, passage\},
``\emph{\{difficulty\}}'' $\in$ \{elementary school, high school, college\}.}
\label{tab:app_sts}
\end{table*}

\begin{table*}[ht]
\centering
\small{\begin{tabular}{l}
\hline
\begin{tabular}[c]{@{}p{0.97\linewidth}@{}}Write a \{unit\} triple with one \{unit\} in \{src\_lang\} and two \{unit\}s in \{tgt\_lang\} with varying translation qualities in JSON format. \\ \\ The triple is denotes as ("S1", "S2", "S3"). The translation quality score ranges from 1 to 5, with higher scores are better.\\ \\ Please adhere to the following guidelines:\\ - The values of "S1" is a string in \{src\_lang\}, the value of "S2" and "S3" are strings in \{tgt\_lang\}.\\ - There should be some word overlaps between "S2" and "S3".\\ - The translation quality score of "S2" with respect to "S1" should be \{high\_score\}.\\ - The translation quality score of "S3" with respect to "S1" should be \{low\_score\}.\\ - "S3" should be grammatical and fluent, but contain some keyword or number translation errors, or miss some information, or contain some redundant information.\\ - "S1" requires \{difficulty\} level education to understand and should be diverse in terms of topic and length.\\ \\ Your output must always be a JSON object only with three keys "S1", "S2" and "S3", do not explain yourself or output anything else. Be creative!\end{tabular} \\ \hline
\end{tabular}}
\caption{Prompt template for bitext retrieval.
For placeholders,
``\emph{\{high\_score\}}'' $\in$ \{4, 4.5, 5\},
``\emph{\{low\_score\}}'' $\in$ \{1.5, 2, 2.5\},
``\emph{\{unit\}}'' $\in$ \{sentence, phrase, passage\},
``\emph{\{difficulty\}}'' $\in$ \{elementary school, high school, college\}.}
\label{tab:app_bitext}
\end{table*}

\section{Prompts for Synthetic Data Generation} \label{sec:app_prompts}
For asymmetric tasks,
we list the four prompt templates in Table~\ref{tab:app_short_long}, ~\ref{tab:app_long_short}, ~\ref{tab:app_short_short}, and ~\ref{tab:app_long_long}.
For symmetric tasks,
the prompts templates are available in Table~\ref{tab:app_sts} and ~\ref{tab:app_bitext}.
To generate multilingual data,
we sample the value of ``\emph{\{language\}}'' from the language list of XLM-R~\citep{conneau2020unsupervised}
with higher probability for high-resource languages.
When prompting GPT-4/3.5,
we set the sampling temperature to $1.0$ and the top-$p$ hyperparameter to $1.0$,
which is higher than the default setting to encourage more diversity.

\section{Instructions for Training and Evaluation} \label{sec:app_instructions}

\begin{table*}[ht]
\centering
\scalebox{0.8}{\begin{tabular}{ll}
\hline
Dataset & Instruction \\ \hline
ELI5 & \begin{tabular}[c]{@{}p{0.85\linewidth}@{}} Provided a user question, retrieve the highest voted answers on Reddit ELI5 forum \end{tabular}  \\
HotpotQA & Given a multi-hop question, retrieve documents that can help answer the question  \\
FEVER & Given a claim, retrieve documents that support or refute the claim  \\
\begin{tabular}[c]{@{}p{0.3\linewidth}@{}}MIRACL / MrTyDi / NQ \\/ SQuAD / TriviaQA\end{tabular} & \begin{tabular}[c]{@{}p{0.85\linewidth}@{}} Given a question, retrieve Wikipedia passages that answer the question \\ Retrieve Wikipedia passages that answer the question \end{tabular}  \\
NLI & \begin{tabular}[c]{@{}p{0.85\linewidth}@{}} Given a premise, retrieve a hypothesis that is entailed by the premise \\ Retrieve semantically similar text \end{tabular}  \\
MS-MARCO & \begin{tabular}[c]{@{}p{0.85\linewidth}@{}} Given a web search query, retrieve relevant passages that answer the query \\ Given a web search query, retrieve relevant documents that answer the query \end{tabular}  \\
Quora Duplicates & \begin{tabular}[c]{@{}p{0.85\linewidth}@{}} Given a question, retrieve questions that are semantically equivalent to the given question \\ Find questions that have the same meaning as the input question \end{tabular}  \\
DuReader / T2Ranking & Given a Chinese search query, retrieve web passages that answer the question  \\ \hline
\end{tabular}}
\caption{Instructions for each training dataset.}
\label{tab:app_train_instructions}
\end{table*}

\begin{table*}[ht]
\centering
\scalebox{0.8}{\begin{tabular}{ll}
\hline
Task Name & Instruction \\ \hline
AmazonCounterfactualClassif. & \begin{tabular}[c]{@{}p{0.85\linewidth}@{}} Classify a given Amazon customer review text as either counterfactual or not-counterfactual\end{tabular}  \\
AmazonPolarityClassification & Classify Amazon reviews into positive or negative sentiment  \\
AmazonReviewsClassification & Classify the given Amazon review into its appropriate rating category  \\
Banking77Classification & Given a online banking query, find the corresponding intents  \\
EmotionClassification & \begin{tabular}[c]{@{}p{0.85\linewidth}@{}} Classify the emotion expressed in the given Twitter message into one of the six emotions: anger, fear, joy, love, sadness, and surprise \end{tabular}  \\
ImdbClassification & Classify the sentiment expressed in the given movie review text from the IMDB dataset  \\
MassiveIntentClassification & Given a user utterance as query, find the user intents  \\
MassiveScenarioClassification & Given a user utterance as query, find the user scenarios  \\
MTOPDomainClassification & Classify the intent domain of the given utterance in task-oriented conversation  \\
MTOPIntentClassification & Classify the intent of the given utterance in task-oriented conversation  \\
ToxicConversationsClassif. & Classify the given comments as either toxic or not toxic  \\
TweetSentimentClassification & Classify the sentiment of a given tweet as either positive, negative, or neutral  \\
ArxivClusteringP2P & \begin{tabular}[c]{@{}p{0.85\linewidth}@{}} Identify the main and secondary category of Arxiv papers based on the titles and abstracts \end{tabular}  \\
ArxivClusteringS2S & Identify the main and secondary category of Arxiv papers based on the titles  \\
BiorxivClusteringP2P & Identify the main category of Biorxiv papers based on the titles and abstracts  \\
BiorxivClusteringS2S & Identify the main category of Biorxiv papers based on the titles  \\
MedrxivClusteringP2P & Identify the main category of Medrxiv papers based on the titles and abstracts  \\
MedrxivClusteringS2S & Identify the main category of Medrxiv papers based on the titles  \\
RedditClustering & Identify the topic or theme of Reddit posts based on the titles  \\
RedditClusteringP2P & Identify the topic or theme of Reddit posts based on the titles and posts  \\
StackExchangeClustering & Identify the topic or theme of StackExchange posts based on the titles  \\
StackExchangeClusteringP2P & Identify the topic or theme of StackExchange posts based on the given paragraphs  \\
TwentyNewsgroupsClustering & Identify the topic or theme of the given news articles  \\
SprintDuplicateQuestions & Retrieve duplicate questions from Sprint forum  \\
TwitterSemEval2015 & Retrieve tweets that are semantically similar to the given tweet  \\
TwitterURLCorpus & Retrieve tweets that are semantically similar to the given tweet  \\
AskUbuntuDupQuestions & Retrieve duplicate questions from AskUbuntu forum  \\
MindSmallReranking & Retrieve relevant news articles based on user browsing history  \\
SciDocsRR & Given a title of a scientific paper, retrieve the titles of other relevant papers  \\
StackOverflowDupQuestions & Retrieve duplicate questions from StackOverflow forum  \\
ArguAna & Given a claim, find documents that refute the claim  \\
ClimateFEVER & Given a claim about climate change, retrieve documents that support or refute the claim  \\
CQADupstackRetrieval & \begin{tabular}[c]{@{}p{0.85\linewidth}@{}} Given a question, retrieve detailed question descriptions from Stackexchange that are duplicates to the given question\end{tabular}  \\
DBPedia & Given a query, retrieve relevant entity descriptions from DBPedia  \\
FEVER & Given a claim, retrieve documents that support or refute the claim  \\
FiQA2018 & Given a financial question, retrieve user replies that best answer the question  \\
HotpotQA & Given a multi-hop question, retrieve documents that can help answer the question  \\
MSMARCO & Given a web search query, retrieve relevant passages that answer the query  \\
NFCorpus & Given a question, retrieve relevant documents that best answer the question  \\
NQ & Given a question, retrieve Wikipedia passages that answer the question  \\
QuoraRetrieval & \begin{tabular}[c]{@{}p{0.85\linewidth}@{}}Given a question, retrieve questions that are semantically equivalent to the given question\end{tabular}  \\
SCIDOCS & Given a scientific paper title, retrieve paper abstracts that are cited by the given paper  \\
SciFact & Given a scientific claim, retrieve documents that support or refute the claim  \\
Touche2020 & Given a question, retrieve detailed and persuasive arguments that answer the question  \\
TRECCOVID & Given a query on COVID-19, retrieve documents that answer the query  \\
STS* & Retrieve semantically similar text.  \\
BUCC/Tatoeba & Retrieve parallel sentences.  \\
SummEval & Given a news summary, retrieve other semantically similar summaries  \\ \hline
\end{tabular}}
\caption{Instructions used for evaluation on the MTEB benchmark.
``STS*'' indicates we use the same instructions for all the STS tasks.}
\label{tab:app_eval_instructions}
\end{table*}

We manually write instructions for training datasets,
as listed in Table~\ref{tab:app_train_instructions}.
For evaluation datasets,
the instructions are listed in Table~\ref{tab:app_eval_instructions}.

\begin{table*}[ht]
\centering
\small{\begin{tabular}{l}
\hline
\begin{tabular}[c]{@{}p{0.97\linewidth}@{}}\textbf{Task group:} short-long matching\end{tabular} \\ \hline
\begin{tabular}[c]{@{}p{0.97\linewidth}@{}}\textbf{Task definition:} The query is a name of a computer software or app, find documents that offer tutorials, guides or user experiences about the software or app.\end{tabular} \\ \hline
\begin{tabular}[c]{@{}p{0.97\linewidth}@{}}\textbf{Generated data:} \{\\  "user\_query": "How to use Microsoft Power BI for data analysis",\\  "positive\_document": "Microsoft Power BI is a sophisticated tool that requires time and practice to master. In this tutorial, we'll show you how to navigate Power BI and perform basic data analysis. First, launch the application and log in with a Microsoft account. Next, create a new project using the Wizard on the left-hand toolbar. Import the dataset you wish to work with by clicking on 'Get Data'. Power BI can handle a variety of file formats, including Excel and CSV. Load your data, then use a sequence of drag-and-drop operations to generate charts and (if applicable) maps. If you are confused at any point, consider using the 'Ask a Question' feature for clarification.",\\  "hard\_negative\_document": "Excel is an incredibly powerful tool for managing and analyzing large amounts of data. Our tutorial series focuses on how you can achieve maximum efficiency with Excel. In this guide, we’ll walk you through transferring data from your computer to an Excel spreadsheet as well as colour-coding cells to match various criteria. We’ll also explore VLOOKUP, conditional formatting, and pivot tables, essentially transforming you into a spreadsheet wizard. Once you’ve mastered these techniques, consider joining us for our Advanced Data Analysis training to branch out even further."\\ \}\end{tabular} \\ \hline \hline
\begin{tabular}[c]{@{}p{0.97\linewidth}@{}}\textbf{Task group:} long-short matching \end{tabular} \\ \hline
\begin{tabular}[c]{@{}p{0.97\linewidth}@{}}\textbf{Task definition:} Identifying severity level of customer complaints in support tickets \end{tabular} \\ \hline
\begin{tabular}[c]{@{}p{0.97\linewidth}@{}}\textbf{Generated data:} \{\\  "input\_text": "I am writing to express my intense dissatisfaction with one of your products, a TV that has stopped functioning only a month after purchase. This situation yields less satisfaction to me and speaks voluminously about your quality control procedures in assembly lines. I hope this troubling issue etches into your improvement list for invoking earnest attention.",\\  "label": "High Severity",\\  "misleading\_label": "Low Severity"\\ \} \end{tabular} \\ \hline \hline
\begin{tabular}[c]{@{}p{0.97\linewidth}@{}}\textbf{Task group:} short-short matching \end{tabular} \\ \hline
\begin{tabular}[c]{@{}p{0.97\linewidth}@{}}\textbf{Task definition:} Provided a movie quote, find the movie title in which it is said. \end{tabular} \\ \hline
\begin{tabular}[c]{@{}p{0.97\linewidth}@{}}\textbf{Generated data:} \{\\  "input": "I'm going to make him an offer he can't refuse.",\\  "positive\_document": "The Godfather"\\ \} \end{tabular} \\ \hline \hline
\begin{tabular}[c]{@{}p{0.97\linewidth}@{}}\textbf{Task group:} long-long matching \end{tabular} \\ \hline
\begin{tabular}[c]{@{}p{0.97\linewidth}@{}}\textbf{Task definition:} Provided a legal brief, retrieve other legal documents presenting similar legal arguments. \end{tabular} \\ \hline
\begin{tabular}[c]{@{}p{0.97\linewidth}@{}}\textbf{Generated data:} \{\\  "input": "In the united states district court for the eastern district of pennsylvania. Plaintiff John Doe brings this action against Defendants ABC Corporation and XYZ Inc., alleging negligence which resulted in his injuries. The Plaintiff alleges ABC Corporation negligently designed, manufactured, and marketed a specific product. The Plaintiff also asserts XYZ Inc. negligently failed to inspect the same product, leading to harmful exposure. Plaintiff respectfully requests this honorable court to hold the defendants liable and award damages. All parties are diverse for purposes of 28 U.S.C. § 1332.",\\ "positive\_document": "Before the Superior Court of California, County of Los Angeles. In the matter of the lawsuit initiated by the Plaintiff Jane Smith against the Defendants PQR Industries and LMN Enterprises, charging they negligently designed, produced, and advertised a product that directly caused harm to her. Furthermore, the Plaintiff has claimed that LMN Enterprises negligently fell short in inspecting the product, leading to unsafe usage. The plaintiff hereby requests this esteemed court to declare the defendants guilty and demands compensation for the damages suffered. According to 28 U.S.C. § 1332, complete diversity exists between the parties to this lawsuit."\\ \} \end{tabular} \\ \hline \hline
\begin{tabular}[c]{@{}p{0.97\linewidth}@{}}\textbf{Task group:} bitext matching \end{tabular} \\ \hline
\begin{tabular}[c]{@{}p{0.97\linewidth}@{}}\textbf{Task definition:} Retrieve parallel sentences. \end{tabular} \\ \hline
\begin{tabular}[c]{@{}p{0.97\linewidth}@{}}\textbf{Generated data:} \{\\  "S1": "Men bugun uzumli tort yasashni rejalashtirdim.",\\  "S2": "I have planned to make a grape cake today.",\\  "S3": "I am going to bake a strawberry pie today."\\ \} \end{tabular} \\ \hline \hline
\begin{tabular}[c]{@{}p{0.97\linewidth}@{}}\textbf{Task group:} monolingual STS \end{tabular} \\ \hline
\begin{tabular}[c]{@{}p{0.97\linewidth}@{}}\textbf{Task definition:} Retrieve semantically similar text. \end{tabular} \\ \hline
\begin{tabular}[c]{@{}p{0.97\linewidth}@{}}\textbf{Generated data:} \{\\  "S1": "Tom loves to ride his blue bicycle in the park every morning.",\\  "S2": "Every morning, Tom enjoys riding his blue bike in the park.",\\  "S3": "Tom takes his blue pen to school every day."\\ \} \end{tabular} \\ \hline \hline
\end{tabular}}
\caption{Random samples for each subgroup of the synthetic data.}
\label{tab:app_synthetic_data_examples}
\end{table*}

\begin{table*}[ht]
\centering
\scalebox{0.8}{\begin{tabular}{lcccc}
\hline
Dataset & w/ synthetic only & w/ synthetic + msmarco & w/o synthetic data & full data \\ \hline
BIOSSES & 84.2 & 81.0 & 85.4 & 85.5 \\
SICK-R & 78.6 & 78.5 & 81.7 & 82.6 \\
STS12 & 75.8 & 74.7 & 77.9 & 79.7 \\
STS13 & 84.3 & 85.3 & 88.0 & 88.4 \\
STS14 & 80.9 & 81.2 & 83.7 & 84.5 \\
STS15 & 86.2 & 86.8 & 89.5 & 90.4 \\
STS16 & 85.0 & 85.3 & 86.5 & 87.7 \\
STS17 & 87.3 & 87.7 & 91.0 & 91.8 \\
STS22 & 66.0 & 67.1 & 66.2 & 67.0 \\
STSBenchmark & 83.5 & 84.0 & 87.8 & 88.6 \\
SummEval & 31.9 & 32.7 & 31.9 & 31.4 \\
SprintDuplicateQuestions & 93.5 & 95.8 & 96.0 & 95.7 \\
TwitterSemEval2015 & 78.0 & 78.5 & 81.7 & 81.6 \\
TwitterURLCorpus & 86.5 & 86.9 & 87.7 & 87.8 \\
AmazonCounterfactualClass. & 79.6 & 79.9 & 77.2 & 78.7 \\
AmazonPolarityClassification & 95.8 & 95.9 & 93.9 & 95.9 \\
AmazonReviewsClassification & 56.9 & 55.5 & 48.2 & 55.8 \\
Banking77Classification & 86.2 & 87.0 & 88.8 & 88.2 \\
EmotionClassification & 49.2 & 47.6 & 51.0 & 49.8 \\
ImdbClassification & 94.8 & 94.9 & 89.0 & 94.8 \\
MassiveIntentClassification & 79.8 & 79.9 & 79.6 & 80.6 \\
MassiveScenarioClassification & 81.7 & 82.4 & 82.3 & 82.4 \\
MTOPDomainClassification & 95.6 & 95.9 & 95.7 & 96.1 \\
MTOPIntentClassification & 84.9 & 85.9 & 83.4 & 86.1 \\
ToxicConversationsClassification & 70.2 & 70.8 & 70.9 & 69.6 \\
TweetSentimentExtractionClass. & 63.5 & 63.4 & 61.6 & 63.7 \\
AskUbuntuDupQuestions & 64.3 & 65.3 & 67.4 & 67.0 \\
MindSmallReranking & 33.1 & 32.8 & 32.5 & 32.6 \\
SciDocsRR & 86.0 & 86.0 & 85.7 & 86.3 \\
StackOverflowDupQuestions & 52.5 & 53.7 & 55.9 & 54.9 \\
ArxivClusteringP2P & 51.4 & 51.2 & 47.8 & 50.5 \\
ArxivClusteringS2S & 46.5 & 44.9 & 44.6 & 45.5 \\
BiorxivClusteringP2P & 44.5 & 43.3 & 36.9 & 43.5 \\
BiorxivClusteringS2S & 40.9 & 40.1 & 37.0 & 40.2 \\
MedrxivClusteringP2P & 40.5 & 39.9 & 32.6 & 38.2 \\
MedrxivClusteringS2S & 38.0 & 37.9 & 32.8 & 37.5 \\
RedditClustering & 56.3 & 55.9 & 63.1 & 57.7 \\
RedditClusteringP2P & 66.3 & 64.8 & 66.4 & 66.5 \\
StackExchangeClustering & 72.9 & 72.7 & 74.5 & 73.1 \\
StackExchangeClusteringP2P & 46.1 & 45.6 & 34.3 & 45.9 \\
TwentyNewsgroupsClustering & 52.2 & 52.5 & 55.6 & 54.3 \\
ArguAna & 52.2 & 42.7 & 62.5 & 61.9 \\
ClimateFEVER & 21.1 & 28.8 & 25.2 & 38.4 \\
CQADupstackAndroidRetrieval & 40.8 & 36.0 & 44.5 & 43.0 \\
DBPedia & 42.0 & 43.7 & 47.7 & 48.9 \\
FEVER & 72.5 & 83.5 & 73.1 & 87.8 \\
FiQA2018 & 38.1 & 48.4 & 54.5 & 56.6 \\
HotpotQA & 48.1 & 64.0 & 75.6 & 75.7 \\
MSMARCO & 25.7 & 45.0 & 42.9 & 43.1 \\
NFCorpus & 35.5 & 40.0 & 35.3 & 38.6 \\
NQ & 53.3 & 63.5 & 57.3 & 63.5 \\
QuoraRetrieval & 75.0 & 79.5 & 89.5 & 89.6 \\
SCIDOCS & 20.6 & 15.8 & 19.0 & 16.3 \\
SciFact & 71.5 & 71.9 & 74.7 & 76.4 \\
Touche2020 & 25.4 & 32.5 & 19.1 & 26.4 \\
TRECCOVID & 82.3 & 87.3 & 70.8 & 87.2 \\ \hline
Average & 63.1 & 64.5 & 64.6 & \textbf{66.6} \\ \hline
\end{tabular}}
\caption{Results for each dataset in the MTEB benchmark.
The evaluation metrics and detailed baseline results are available in the original paper~\citep{muennighoff2023mteb}.}
\label{tab:app_full_results}
\end{table*}

\end{document}